\documentclass[sigconf]{acmart} 

\AtBeginDocument{%
  }

\setcopyright{acmlicensed}
\copyrightyear{2026}
\acmYear{2026}
\acmDOI{XXXXXXX.XXXXXXX}
\acmConference[Under review for KDD '26]{Make sure to enter the correct
  conference title from your rights confirmation email}{August 09--13,
  2026}{Jeju, Korea}
\acmISBN{978-1-4503-XXXX-X/2018/06}

\usepackage{subcaption}
\usepackage{multirow}

\begin{document}

\title{Positive-Unlabelled Active Learning to Curate a Dataset for Orca Resident Interpretation}

\author{Bret Nestor}
\authornote{Work completed while at the University of Toronto and the University of Washington}
\authornote{Corresponding author}
\email{bret.nestor@ubc.ca}
\orcid{1234-5678-9012}
\affiliation{%
  \institution{Translacean Research Foundation}
  \city{Vancouver}
  \country{Canada}
}
\affiliation{%
  \institution{University of British Columbia}
  \city{Vancouver}
  \country{Canada}
}

\author{Bohan Yao}
\affiliation{%
  \institution{University of Washington}
  \city{Seattle}
  \country{United States}}

\author{Jasmine Moore}
\affiliation{%
  \institution{University of Calgary}
  \city{Calgary}
  \country{Canada}
}

\author{Jasper Kanes}
\affiliation{%
 \institution{Ocean Networks Canada}
 \city{Victoria}
 \country{Canada}}

\renewcommand{\shortauthors}{Nestor et al.}

\begin{abstract}
  This work presents the largest curation of Southern Resident Killer Whale (SRKW) acoustic data to date, also containing other marine mammals in their environment. We systematically search all available public archival hydrophone data within the SRKW habitat (over 30 years of audio data). The search consists of a weakly-supervised, positive-unlabelled, active learning strategy to identify all instances of marine mammals. The resulting transformer-based presence or absence classifiers outperform state-of-the-art classifiers on 3 of 4 expert-annotated datasets in terms of accuracy and energy efficiency. The fleet of WHISPER detection models range from 0.58 (0.48-0.67) AUROC with WHISPER-tiny to 0.77 (0.63-0.93) with WHISPER-large-v3. Our multiclass species classifier obtains a top-1 accuracy of 53.2\% (11 train classes, 4 test classes) and our ecotype classifier obtains a top-1 accuracy of 33.6\% (4 train classes, 5 test classes) on the DCLDE-2026 dataset.
  
  We yield 919 hours of SRKW data, 230 hours of Bigg's orca data, 1374 hours of orca data from unlabelled ecotypes, 1501 hours of humpback data, 88 hours of sea lion data, 246 hours of pacific white-sided dolphin data, and over 784 hours of unspecified marine mammal data. This SRKW dataset is larger than DCLDE-2026, Ocean Networks Canada, and OrcaSound combined. The curated species labels are available under CC-BY 4.0 license, and the corresponding audio data  are available under the licenses of the original owners. The comprehensive nature of this dataset makes it suitable for unsupervised machine translation, habitat usage surveys, and conservation endeavours for this critically endangered ecotype.
  
\end{abstract}



\received{9 February 2026}

\maketitle

\section{Introduction}

The southern resident killer whales (SRKWs) are an ecotype of Orcinus orca who inhabit the waters of the northeastern pacific. They are designated as endangered by both the Department of Fisheries and Oceans Canada~\cite{Canada_endangered} and the U.S. Fish and Wildlife Service~\cite{US_endangered}. The International Union for the Conservation of Nature (IUCN) lists orcas' endangered status as "data deficient" because they are comprised of diverse ecotypes~\cite{iucn}, though recent work has proposed that Northeast Pacific ecotypes should be classified as distinct species~\cite{doi:10.1098/rsos.231368}. With this distinction, SRKWs who diverged from other lineages approximately 100,000 years ago, could also receive threatened status under the IUCN~\cite{doi:10.1098/rsos.231368}. 

The critical decrease in SRKW population was caused by extensive capture for entertainment, compounded by a failure to rebound due to decreased prey availability owing to dam construction, increased contaminants, and increased noise pollution~\cite{Lacy2017}. They face direct threats from human activity, such as ship strikes~\cite{10.1371/journal.pone.0242505} and ship noise~\cite{vagle2021vessel}, which reduces their ability to echo-locate prey~\cite{https://doi.org/10.1111/gcb.17490}. They also face threats indirectly caused by human activity, such as a lack of large Chinook salmon prey availability~\cite{10.1371/journal.pone.0247031,Oke2020}

In the Northeastern Pacific the SRKWs diet constitutes fish, whereas Bigg's orcas are mammal-eating~\cite{Bigg1982AnAO}. Due to their choice of prey, resident orcas are more vocal than the mammal-eating Bigg's ecotype~\citep{DEECKE2005395}. The SRKW socialisation and echolocation make them easier to detect acoustically, as compared to other cetaceans. As their habitat coincides with the heavily monitored Salish Sea and Northeast Pacific Ocean, we hypothesise that there are ample instances of SRKWs within unsearched archival datasets.

\textbf{The goal of this work is to collect all archival instances of SRKWs to build a Dataset for Orca Resident Interpretation (DORI).} Accomplishing this requires 1) using effective classifiers to discover instances of SRKW data from unlabelled archival data, 2) developing effective species and ecotype labelling techniques, and 3) labelling data.

We develop competitive marine mammal detection classifiers, and species classifiers, resulting in the largest curation of marine mammal acoustic data to date ~\footnote{Data available at ~\url{https://huggingface.co/collections/DORI-SRKW/dori}}. From a set of more than 30 years (260,000 hours) of archival data, we discover 919 hours of SRKW data.

\subsection{Passive Acoustic Monitoring}

Passive acoustic monitoring (PAM) involves setting up a recording device in a region of interest, and permitting it to record indiscriminately without human oversight~\cite{f9449875-5719-3b43-98fb-c569297e2425}. Data is later analysed to determine the presence of sound-producing species. PAM datasets are unique because they are recorded in-the-wild without confounding research vessel interference~\cite{WILLIAMS2006301,D2009}. One of the unique challenges of passive acoustic data is that it generates a tremendous amount of data, yet only a small fraction of it involves the species of interest. This results in datasets with extreme class imbalance. Historically, PAM was reviewed by human annotators~\cite{Tzanetakis}, but algorithms have been developed to preprocess and filter acoustic data at scale, and to automatically detect species presence within these large-scale datasets~\cite{pamguard, birdnet}. Audio is often record audio with a duty-cycle to reduce labelling effort or storage requirements in remote locations. Though duty cycling could potentially cause missed identifications~\cite{10.1121/1.4816552} or inaccurate durations~\cite{10.1121/10.0009752}.

Human-annotation would be infeasible the scale required to search this archive~\cite{rand2022effects}. The most comprehensive approach to date has been Orchive, a digitalization of 20,000 hours of acoustic data from Orcalab, in northern resident killer whale habitat~\cite{Tzanetakis}. SRKW acoustic surveys been fragmented. From 2006-2011, several hydrophone deployments along the United States coast detected 131 SRKW events~\cite{10.1121/1.4821206}. From 2009-2011, 175 days of monitoring contained SRKWs at Swiftsure Bank~\cite{Riera2019}. From 2015-2017, 96 days of SRKW events were detected in the winter season in the northern Salish Sea~\cite{10.3389/fmars.2023.1204908}. In 2018, 46 days with SRKW events were detected off of Lime Kiln~\cite{10.1121/10.0009752}. Between 2018-2020, 605 days contained SRKW presence at Swiftsure Bank~\cite{Riera2019}. These studies included duty cycles and intermittent deployment. Cumulatively, SRKWs were detected on 1053 days. These acoustic data are not available publicly. However, during these same periods, there are publicly available (CC-BY-NC-SA 4.0) recordings from OrcaSound, amounting to 122 whale days from 2018-2022.


\subsection{Active Learning and Positive-Unlabelled Learning}
Active learning is a process that reduces labelling effort by prioritising samples that are informative. Typically this is done by sorting samples according to their uncertainty, diversity, or model influence~\cite{10537213}. A key aspect of active learning is iterating between model retraining, and labelling unlabelled samples that would maximally benefit the model. When high entropy samples are selected for active learning, the resulting model outperforms a model trained on the same quantity of randomly labelled data~\cite{active}. In the binary classification case this corresponds to a logit probability closest to 0.5~\cite{active_orig}. There is a natural tension between exploring within-class diversity and samples far from class distributions. Modern algorithms seek to strike a balance between these, especially when classes are imbalanced~\cite{9093475}, heterogeneous~\cite{active}, or unknown (known as open-set annotation)~\cite{Ning_2022_CVPR}. Other works seek to be task-agnostic by predicting the model's loss from the covariates in order to sample under-represented and highly influential samples~\cite{learningloss}. In contrast to PAM approaches, which duty-cycle the unlabelled set to have an unbiased measure of the data, active learning takes a biased selection of the data to uncover the maximum model performance per labelling effort.

Under class-conditional mislabelling, simple models are noise tolerant~\cite{NIPS2013_3871bd64}. This means that when candidates are mislabelled, the classifier trained on these mislabelled samples can still approximate the performance of a classifier trained on correctly labelled samples. In the most extreme case of positive-unlabelled learning (where noise is only applied to randomly flip some of the positive labels to negatives/unlabelled) the loss function for a weighted regression problem serves as a surrogate for the true underlying loss~\cite{NIPS2017_7cce53cf}. While it relies on knowing the class proportions, this is only a single hyperparameter that is tuned in cross validation~\cite{NIPS2013_3871bd64}. The convergence rate of PU learning has been derived to be:
\[ \begin{cases} 
      \mathcal{O}(\frac{V}{n\cdot e_m \cdot h}) & h \geq \sqrt{\frac{V}{n\cdot e_m}} \\
      \mathcal{O}(\sqrt{V/(n\cdot e_m )}) & h < \sqrt{\frac{V}{n\cdot e_m}}  \\
   \end{cases}
\]

Where $V$ is the Vapnik-Chervonenkis dimension, $n$ is the number of samples, $e_m$ is the propensity of a positive sample to be labelled under the selected completely-at-random (SCAR) assumption, and $h$ is Massart margin that quantifies the difficulty of classifying the data~\cite{JMLR:v24:22-067}. A lower Massart margin, $h$, indicates a harder to classify example. Fortunately, in the active learning setting $e_m$ approaches 1 as labelling progresses. This drives the bound towards the standard classification bound. For a difficult to classify problem, or for a large pretrained model, having unlabelled positive samples in your dataset incurs virtually no penalty as compared to a standard training setup when the negative instances vastly outnumber the positive instances. 

These bounds show that models with sufficient data can lower their empirical risk on \textit{I.I.D.} test data irrespective of architecture choices. We use transformer-based models that are universal function approximators for sequential data problems~\cite{Yun2020Are}.

Knowing that performant models can be uncovered using noisy labels is fortunate for passive acoustic monitoring, since we can rely on non-experts. The challenge for machine learning models in this problem setting is that the SCAR assumption used to derive the bounds may be broken by selectively identifying high SNR samples. Systematic archive searches require amateur (noisy) labellers, since the quantity of available archival data grows faster than experts can process them. In a related work investigating the ability of citizen scientists to annotate deep-sea images, the citizen scientists positively identified 50-65\% of expert annotated positive images, while maintaining a true positive rate of 36-46\%~\cite{10.1371/journal.pone.0218086}. This enriched sensitivity reduces expert burden so long as it is above the ambient positivity rate of the data. While these capabilities might not transfer from imaging to audio annotation, we hypothesize that amateur listeners can similarly identify high-SNR instances leading to ample labelled marine mammal acoustic events.

We hypothesize that by combining positive-unlabelled learning, amateur labelling, and active-learning, we can recover classifiers that often outperform existing state-of-the art classifiers. The resulting models should be small enough and efficient enough to search the ever growing archives of SRKW habitat.

\section{Methods}
\begin{figure*}[h]
  \centering
  \includegraphics[width=\linewidth]{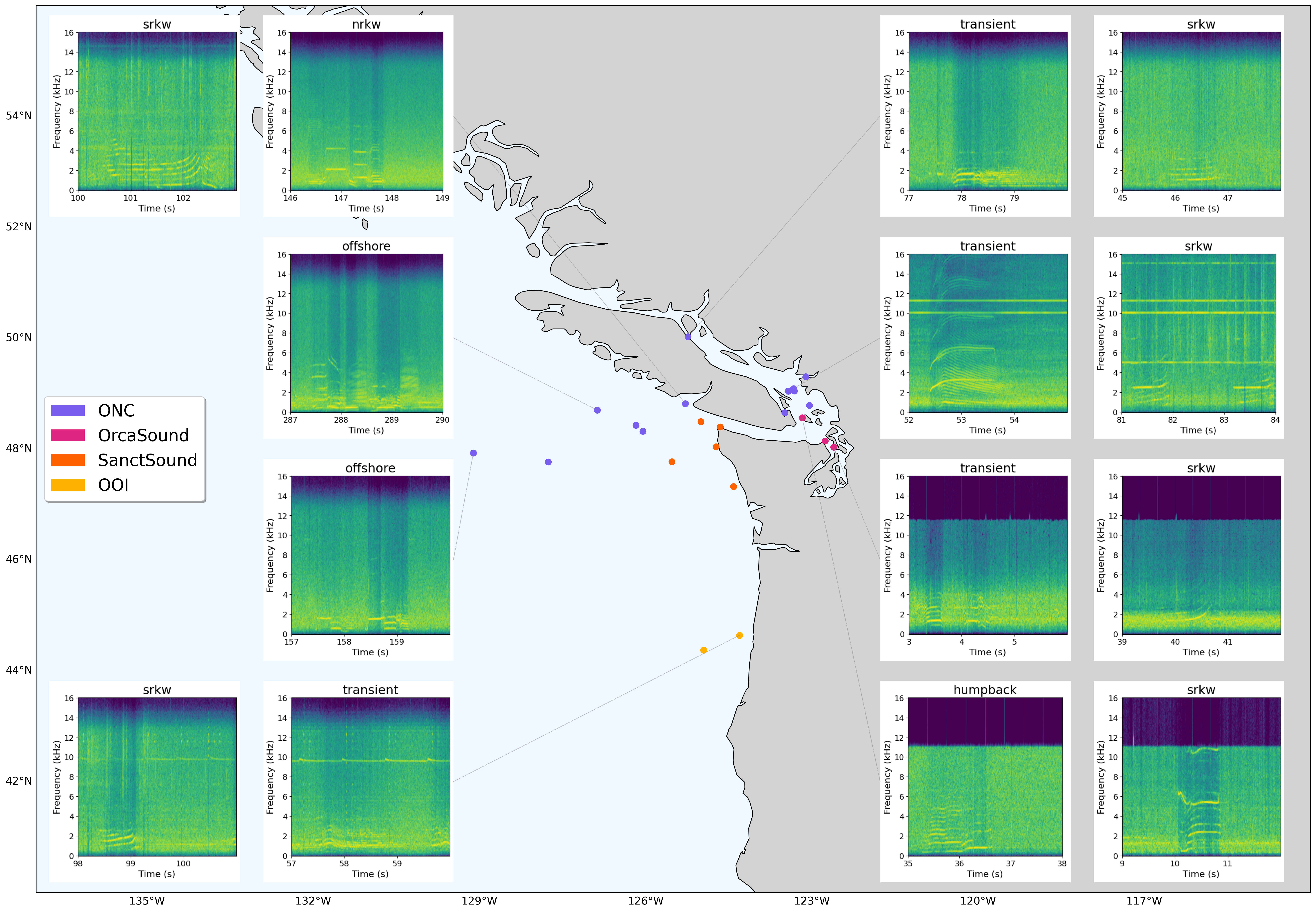}
  \caption{Hydrophone deployment sites where data are collected from.}
  \Description{Hydrophone deployment sites in the Northeast Pacific with spectrogram examples overlaid.}
  \label{fig:schematic}
\end{figure*}

A schematic overview of data sources are given in Figure~\ref{fig:schematic}.

\subsection{Data Sources}
We curate data within the region bounded by 40N to 51N latitude, and -132E to -120E longitude, through the end of 2022. We only explore publicly available hydrophone data under permissive data use licenses or agreements.

\paragraph{Ocean Networks Canada}
The Ocean Networks Canada archival repository was the largest source investigated, spanning 60057 days of hydrophone deployments. Due to storage considerations, we only retained 3143 days of high likelihood hydrophone data for the active learning simulation. These data are distributed under a CC-BY license.

\paragraph{SanctSound}
SanctSound involved temporary hydrophone deployments along the outer Washington coast between 2018 and 2022~\cite{Hatch2024}. We analysed 1788 days of data. Due to storage considerations, we only retained 455 days of the highest probability samples. Machine learning detections of Orcas were provided with annotated start and end time of acoustic bouts. Ecotype determinations were not included. Humpbacks were labelled by day-level presence, rather than in bouts.

\paragraph{Ocean Observatories Initiative}
Ocean Observatories initiative operates hydrophone nodes near the continental shelf and Yaquina Bay~\cite{ooi}. These data have not been thoroughly explored for marine mammal events. The archives contain 10370 days of hydrophone deployments within the specified range, usable under a custom terms\footnote{\url{https://oceanobservatories.org/how-to-use-acknowledge-and-cite-data/}}.

\paragraph{OrcaSound}
OrcaSound is a community collective of self hosted hydrophone nodes and volunteer annotators~\cite{orcasound}. Their data contain human and machine detections through various deployments in the southern Salish Sea. We include 58 days of data from 4 locations from 2018- 2022. These data were selected because they coincided with nearby sightings, so they represent an enriched prevalence.

\paragraph{DCLDE 2026 dataset}
This dataset is curated specifically for the 2026 Biennial Conference and Workshop on Detection, Classification, Localization, and Density Estimation of Marine Mammals using Passive Acoustics (DCLDE)~\cite{Palmer2025}. There are 35 hours of marine mammal audio, including 12.2 hours of killer whale data, and 4.1 hours of SRKW data. It contains several orca ecotypes for ecotype classification problems. Some of the contributed data overlaps with the data from the previously listed sources. For this reason, we intentionally omit OrcaSound and Ocean Networks Canada data when this dataset is used as an external test set.

\paragraph{DEEPAL Fieldwork}
This dataset is collected using a towed hydrophone array, with the intention of classifying audio clips as containing, or not containing orcas~\cite{orcaspot, Bergler2022}. The dataset requires a data use agreement to access.~\footnote{access instructions at https://lme.tf.fau.de/dataset/deepal-fieldwork-data-2017-2018-dlfd/}

\paragraph{FGPD \& PSGCH 20203 datasets}
These datasets are entirely sourced from Ocean Networks Canada, and retain the CC-BY 4.0 license for the data. The FGPD set features 365 5-minute segments of data from a hydrophone node at hahuułi~\cite{oceannetworkscanada2020} (lat 48.8138, lon , -125.2746, depth 95m), and PSGCH contains 31 5-minute segments from August 2023 in the Salish Sea~\cite{oceannetworkscanada2023} (lat 49.0395, lon -123.4252, depth 298). The segments were collected by taking the maximum scored sample for a given day, using logits drawn from several models. This means that the segment-level prevalence of marine mammals is over-representative, but the day-level prevalence of marine mammals is representative. The labels are 1, if a marine mammal is present, and a 0 otherwise (including snapping shrimp, fish grunts). The marine mammals in FGPD include humpbacks, Bigg's transients, NRKW, SRKW, and sea lions. PSGCH contains humpbacks and Bigg's transients. The introduced test sets are chronologically disjoint from our training set, and are labelled by an expert bioacoustician.

Other hydrophone operators are also in the region, such as the British Columbia Hydrophone Network, Raincoast Conservation Foundation, and the Lime Kiln State Park Lighthouse. These sites were not included because they did not have comprehensive coverage, permissive data, or archival data in the study period.

\paragraph{Sightings data}
We obtained records from the Pacific Whale Watch Association (under a limited data use agreement), Orca Networks' archival data~\cite{orcanetwork}, iNaturalist~\cite{inaturalist,gbif}, and prior literature~\cite{shields20232018} to build a thorough profile of sightings. In total, there were 92745 unique sightings, of which 855 were reported within 1 km of a hydrophone.

\subsection{Preprocessing}
Audio files were converted from their original format to a lossless \texttt{flac} encoding for storage purposes using ffmpeg (\url{https://ffmpeg.org/}). If the audio files were longer than 5 minutes, they were also separated into 5-minute long segments at this step. Each segment which was 5-minutes long or less was  resampled to 32 kHz (to maintain a Nyquist frequency of 16 kHz) and processed through a high-pass filter at 1 kHz. The default whisper-tiny parameters were used for mel spectrogram preprocessing (a \textit{chunk\_length} of 30, a \textit{feature\_size} of 80, and a \textit{hop\_length} of 160). 15-second long segments of each file were fed through the 39 million parameter whisper-tiny encoder~\cite{whisper} to obtain a feature set with the dimensions of \textit{audio\_chunks},\textit{hidden\_steps}, \textit{hidden\_size}. Layer normalization was applied to the outputs at this stage~\cite{ba2016layer}. A mean-pooling was applied across the \textit{hidden\_steps} dimension. All dimensions were averaged until we obtained a single vector per file with the dimension: \textit{hidden\_size}. We refer to these per-file vector representations as embeddings throughout the rest of the paper.

The median embedding required 0.000045 kWh of electricity to process~\cite{benoit_courty_2024_11171501}. The consumption of all embeddings was 153.37 kWh, which was conducted in regions with >90\% of electricity sourced from nuclear, hydroelectric, wind, or solar. Less than 10 kg of CO$_2$ was produced while generating the embeddings, which was offset through the Great Bear Forest Carbon Project (\url{https://www.carbonzero.ca}).

\subsection{Weakly Supervised Learning}
The embeddings were given to a standard logistic regression classifier implemented in scikit-learn with an L-2 penalty~\cite{scikit-learn}. We also train a classifier using SGD and a hinge-loss. The purpose of this hinge loss is to evaluate a model training scheme that removes the bias caused by deleting unlikely samples. Positive labels included all "seed" labels, which were ONC labelled data, and all human labelled positive instances at the time of training. Since the positive instances were so rare, samples were randomly selected from the unlabelled data to serve as negative instances during training. The class priors were not adjusted to accommodate the likeliness of the negative instances of being positive. We leave this to future work, as we can adjust the priors based on location, bathymetry, time-of-year, and time-of-day, to inform the priors for each sample. In some active learning strategies definitive negatives (those already labelled by a human) supplemented the negative class. Data were split by time, with audio files recorded before 2021 being relegated to training, those recorded in 2021 used as validation, and those in 2022 as a held-out test set.

\subsection{Active Learning}
We followed several different active learning strategies. This included \textbf{Positive only} sampling where at each step only the top k samples were labelled. Files that were not labelled positive remained in the unlabelled group in case a low SNR sample was missed. Next, we applied a standard \textbf{entropy-based} labelling strategy~\cite{active}. The top-k samples sorted in descending order by entropy were labelled. Samples which were labelled as negatives were kept as guaranteed negatives for future model training iterations. We employ a \textbf{loss-estimate} strategy as described in prior work~\cite{learningloss}. The labelling strategy was similar to that of the entropy strategy, except that samples were sorted by descending order of predicted loss. Two hybrid approaches were conducted, the first being a mixture of the positive and entropy sampling approaches. At each iteration, half of the labelling budget was applied to the positive labelling strategy and half to the entropy labelling strategy. We refer to this approach as ~\textbf{mixed} active learning. The second hybrid approach, ~\textbf{alternating}, would alternate the labelling strategy between the positive labelling and entropy labelling strategies. For these experiments, 500 5-minute long samples were labelled at each labelling step.

We simulate the learning process with 5 randomly seeded models. We demonstrate the experiments using the labels produced in this work, and in a second setting where 30\% of the positive labels are randomly flipped to negative for each new batch of 500 labels.

\subsection{Labelling}
Data were labelled using both the available spectrogram and audio. The labelling strategy used to curate the dataset most closely resembled the "alternating" approach. In addition to purely algorithmic searches, heuristics were applied to try to capture more samples; Every file before and after a marine mammal detection was manually searched, regardless of its likelihood. Hydrophone array deployments were ensured to have the same label for each device to reduce the number of false negatives. Due to data storage capacity, low-probability samples from ONC were periodically deleted and removed from disk throughout the labelling protocol. For this reason, our embeddings are slightly over-representative of the SRKW prevalence.

Species labels were seeded by using previous ONC annotations, and Discovery of Sound in the Sea examples~\cite{dosits}. Ecotype determinations were made by labelling calls according to those recorded in prior literature~\cite{Ford_1984, Souhaut2021-ij}. When available, sightings data were used to clarify species and ecotype labels.

\subsection{Benchmark Models}

As a benchmark, we re-implement the PAMguard Whistle\&Moan detector and \textbf{LDA classifier}~\cite{pamguard, pamguard_techniques}. We also adapt the \textbf{ROCCA random forest classifier}~\cite{rocca} as a detection model. Audio data are loaded and resampled to 32 kHz. Echolocation clicks are removed using a sliding exponential filter:
$$signal_{low} = \frac{1}{\left(1+\frac{signal-mean(signal)}{\gamma*std(signal)}\right)^p}$$
where $p$ and $\gamma$ are hyperparameters. A spectrogram is created from the transformed audio data and converted into decibels. Spectral noise is removed by subtracting a sliding median filter over the spectrogram, introducing a kernel size hyperparameter, $\kappa$. The decaying running-average is subtracted to remove tonal noise, introducing an exponential decay hyperparameter, $\alpha$. Subsequently, Gaussian smoothing is applied to the spectrogram.  The spectrogram is then binarised by thresholding it at a specific decibel value, $\beta$. Pixels are grouped into regions by connecting groups of neighbouring pixels. Regions with fewer than $min\_length$ pixels and fewer than $min\_count$ pixels are omitted. The remaining binary map should highlight only the detections of cetacean whistles and moans. The hyperparameters $\gamma$, $p$, $\alpha$,  $\kappa$, $\beta$, $min\_length$, and $min\_count$ are tuned using a random hyperparameter search. In addition, an optional high-pass filter at 1 kHz is added to the hyperparameter space to remove low frequency noise.

While this detection filter does not constitute a machine learning algorithm, the PAMGuard software~\cite{pamguard,pamguard_techniques} also employs a latent discriminant analysis (LDA) classifier and an implementation of the random forest classifier highlighted in prior work~\cite{rocca}.
For the LDA classifier, features are procured from the binarised spectrogram by cycling through the regions. If a region is of  sufficiently long duration, it is split into slices. A second order polynomial is fit to the binary pixels in each slice. The zeroth, first, and second coefficients of the polynomials are collected for each slice. As a result, for any spectrogram, there is a distribution of these coefficients. Subsequently, the mean, standard deviation, and skew are calculated for the distribution, resulting in nine features per audio file. Though the original purpose of this classifier was to classify various species, we use it also as an enriched presence or absence classifier. LDA classifiers are then hyperparameter tuned for the detection filter hyperparameters as well as slice length.

We employed a similar approach for the ROCCA random forest algorithm~\cite{rocca}. In the original implementation, ROCCA relies upon human-generated whistle contours to create classifications,  which is infeasible on large datasets. Therefore, we opted to use the contours generated by the PAMGuard whistle \& moan detector. In some cases, this could generate over 10,000 contours per 5-minute audio file. For each contour, 57 attributes of the signal were extracted (i.e., starting frequency, slope, harmonics)~\cite{rocca}. Training was conducted using weakly supervised labels (i.e., propagating the label from the entire spectrogram to each contour). For hyperparameter tuning on the validation set, all contours in the spectrogram were ensembled such that the label receiving the most votes became the classification for the spectrogram.

For both of these techniques, it is important to note that the algorithm does not have a consistent computational cost per audio file, but rather depend on the number of contours produced in the whistle \& moan detector.

We also use the ~\textbf{Animal-Spot} library as a benchmark~\cite{Bergler2022, orcaspot}. This model is a ResNet-based classifier~\cite{resnet} that classifies fixed spectrogram segments, as is common in passive acoustic monitoring~\cite{birdnet}. We train the 11.69K parameter ResNet18 on the DEEPAL fieldwork dataset's training split and hyperparameter tune using the validation split. This dataset has two classes, "noise" and "orca". To make the model more competitive in speed and efficiency, we enable mixed precision during inference, and modify the data-loader to load 5-minute increments in parallel and split them into the appropriately formatted spectrograms.

\section{Results}
\subsection{Active Learning}
\begin{figure}[ht]
    \centering
    
    \begin{subfigure}[b]{0.98\linewidth}
      \includegraphics[width=1\linewidth]{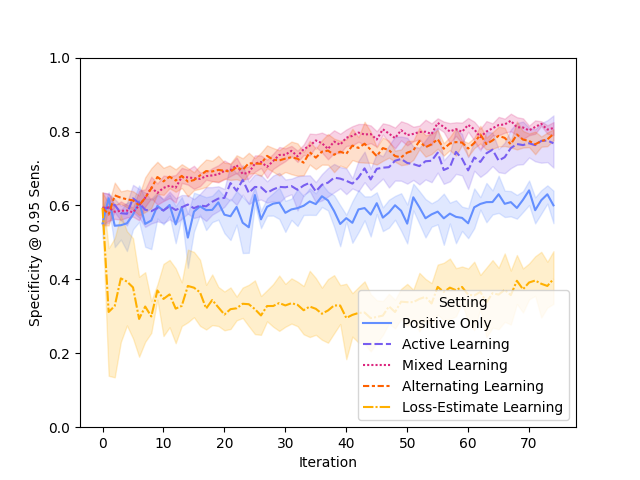}
      \caption{Specificity at 95\% Sensitivity vs labelling iterations.}
      \label{fig:Ng1} 
    \end{subfigure}
    
    \medskip 
    \begin{subfigure}[b]{0.98\linewidth}
      \includegraphics[width=1\linewidth]{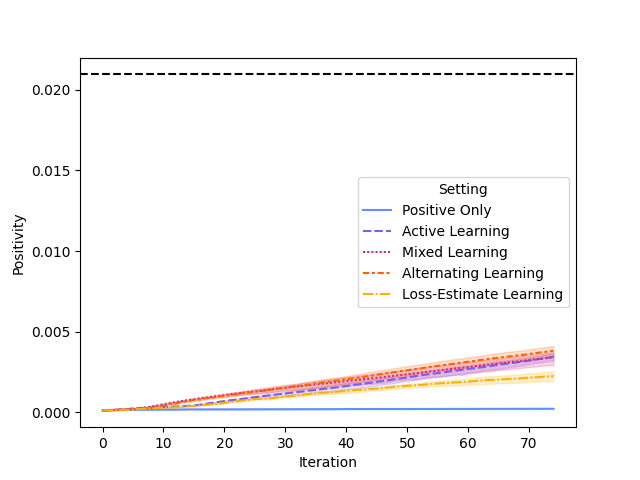}
      \caption{Positivity rate vs. active learning iteration. The dashed line represents the positivity rate for the entire embedding dataset.}
      \label{fig:Ng2}
    \end{subfigure}
    
    \caption[No noise]{%
    Active learning without noisy labels. Mixing positive and active learning helps discover samples with no penalty compared to entropy-only sampling.}
    \label{fig:no_noise}
    \Description{Active learning experiments under no label noise.}

\end{figure}

\begin{figure}[ht]
    \centering
    
    \begin{subfigure}[b]{0.98\linewidth}
      \includegraphics[width=1\linewidth]{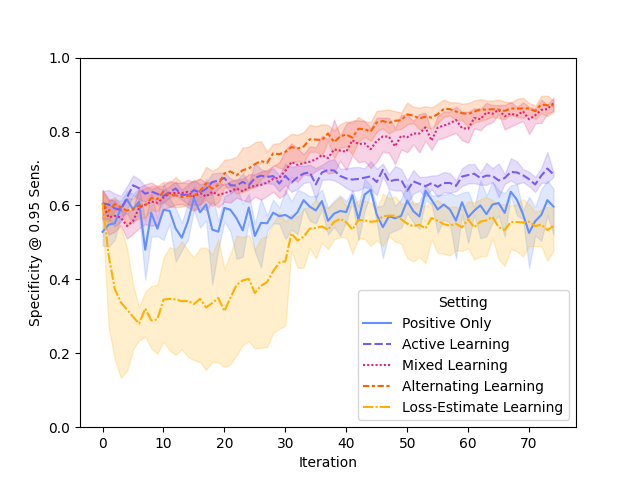}
      \caption{Specificity at 95\% Sensitivity vs labelling iterations.}
      \label{fig:Ng3} 
    \end{subfigure}
    
    \medskip 
    \begin{subfigure}[b]{0.98\linewidth}
      \includegraphics[width=1\linewidth]{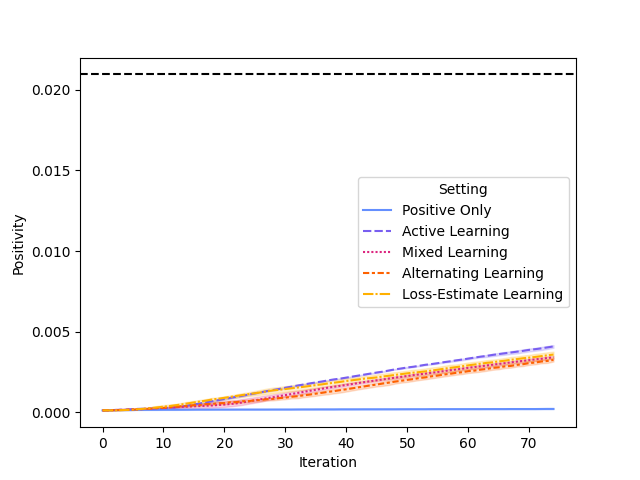}
      \caption{Positivity rate vs. active learning iteration. The dashed line represents the positivity rate for the entire embedding dataset.}
      \label{fig:Ng4}
    \end{subfigure}
    
    \caption[Noise]{%
    Active learning with label noise. At each iteration 30\% of positively labelled samples are assigned flipped to negatives. In this setting, entropy-only sampling slightly outperforms the mixed strategies.}
    \label{fig:noise}
    \Description{Active learning experiments under label noise}

\end{figure}

The positive-unlabelled active learning experiments are demonstrated in Figures~\ref{fig:no_noise}~\&~\ref{fig:noise} for the positive-unlabelled setting on the disjoint testing dataset (2022). There are two objectives of this approach; one is to develop high-performing classifiers to maximise the likelihood of labelling positive instances, and the other objective is to label all possible positive instances .

First, we seek to evaluate the classifier performance across labelling iterations. When trying to construct the best possible classifier (as evaluated with validation and test performance as a function of labelling iterations), the mixed labelling strategies are preferred (Figure~\ref{fig:Ng1}). This is echoed in the second objective where the mixed labelling strategies are slightly more efficient at discovering data (Figure~\ref{fig:Ng2}).




Bioacousticians may be wary of trusting models trained using non-expert labelled data. We simulate scenarios where varying percentages of true positives are missed. Anecdotally, these may be low SNR cases where there is surrounding vessel noise or distant marine mammals. Under these scenarios, we surprisingly observe that the models using noisy data and a mixed or alternating labelling approach ~\textit{outperform} the clean data (Figure~\ref{fig:Ng3}). This exploration comes at an expense. We observe that under the noisy-labelling scenario entropy-based active learning discovers positive instances faster than the other approaches (Figure~\ref{fig:Ng4}).




\subsection{Benchmarking}
Next, we compare the capabilities of our classifiers to the pre-existing techniques. We find that compared to PAMGuard Whistle \& Moan detector + LDA~\cite{pamguard}, ROCCA~\cite{rocca}, and ANIMAL-SPOT~\cite{Bergler2022,orcaspot} the fine-tuned whisper is competitive across several held out test sets (Table~\ref{tab:held_out}). While Animal-Spot attains highest performance on the domain it is trained on, it is unsurprising that the whisper-model, fine-tuned with the most data generalizes well to in-domain and out-of-domain datasets. As model capacity grows, and the number of trainable parameters increases, so did the AUROC on the held-out test sets.

The whisper and ResNet models exceeded real-time detection (Table~\ref{tab:efficiency}). The whisper model was more energy efficient during inference regardless of the hardware. Animal-Spot and whisper-tiny were competitive in processing time on GPU. In 1 day of consumer GPU processing, almost an entire year of 32 kHz PAM data can be processed on a single GPU.

\begin{table*}[h]
\begin{center}
\begin{minipage}{\textwidth}
\caption{Cetacean detection AUROC on several held out test sets. "LR" means logistic regression and "SGD" indicates stochastic gradient descent with a hinge loss on frozen embeddings. "FT" indicates that all weights were unfrozen and updated. * a maximum of 1000 contours are subsampled. }\label{tab:held_out}%

\centering
\begin{tabular}{lrrrr}
\toprule
  Model & Deepal & DCLDE-2026~\cite{Palmer2025} & FGPD & PSGCH.H3 \\
   & ComParE~\cite{orcaspot}  & excl. ONC \& OS & 2023 & 2023/08 \\
\midrule
PAMGaurd LDA\footnote{Algorithm implemented in python to operate across platforms. Hyper-parameter tuning was conducted to outperform the PAMGuard defaults.} & 0.520 & 0.559 & 0.490 & 0.390 \\
ROCCA\footnote{Algorithm implemented in python to operate across platforms. Hyper-parameter tuning was conducted to outperform the PAMGuard defaults.}~\cite{rocca} &  0.467 & 0.533\footnote{Generated contours were randomly subsampled so that only 1000 contours were selected}   & 0.582 & 0.548 \\
ANIMAL-SPOT~\cite{Bergler2022} & 0.648 & 0.466 & 0.529 & 0.348 \\
Whisper-tiny~\cite{whisper} + LR & 0.480 & 0.669 & 0.542 & 0.624 \\
Whisper-tiny~\cite{whisper} + SGD & 0.600 & 0.671 & 0.644 & 0.852 \\
Whisper-tiny~\cite{whisper} + FT & 0.568 & \textbf{0.786} & \textbf{0.678} & 0.957 \\
Whisper-base~\cite{whisper} + FT & \textbf{0.732} & 0.749 & 0.599 & 0.786 \\
Whisper-large-v3~\cite{whisper} + FT & 0.716 & 0.779 & 0.632 & \textbf{0.933} \\
Wav2Vec 2~\cite{NEURIPS2020_92d1e1eb} +FT  & 0.604 & 0.655 & 0.568 & 0.657 \\

\bottomrule
\end{tabular}
\end{minipage}
\end{center}
\end{table*}

\begin{table*}[h]
\begin{center}
\begin{minipage}{\textwidth}
\caption{Computational efficiency of presence/absence classifiers. All benchmarks are conducted on the FGPD and PSGCH.H3. Except for LDA and ROCCA, all settings used 8 concurrent workers. "+S" means that the contours found are subsampled to increase the file processing throughput.}
\label{tab:efficiency}%
\centering
\begin{tabular}{@{}llrrrr@{}}
\toprule
 System & Model & Processing Time & Power Efficiency & RAM & VRAM\\
 & &  (audio:compute) & (day audio/$kWh$) & (MB) & (MB) \\
  & &  $\uparrow$ & $\uparrow$ & $\downarrow$ & $\downarrow$ \\
\midrule
\multirow{1}{*}{CPU} &PAMGuard LDA~\cite{pamguard} & 55.34 & \textbf{46.80} & 6130 & - \\
& ROCCA~\cite{pamguard,rocca} &  0.494  &  0.06 & 16368 & - \\  
& ROCCA\footnote{Generated contours were randomly subsampled so that only 1000 contours were selected}~\cite{pamguard,rocca} &  20.23 &  17.29 & 5771 & - \\  


& Animal-Spot~\cite{Bergler2022} & 25.02 & 7.41 & 6512 & -   \\ 


& Whisper-tiny~\cite{whisper} & \textbf{122.09} & 26.51 & \textbf{2237} & - \\ 



& Whisper-base~\cite{whisper} & 23.38 & 4.95 & 3388 & - \\ 



& Whisper-large-v3~\cite{whisper} & 4.42 & 1.20 & 7394 & - \\ 



& Wav2Vec 2~\cite{NEURIPS2020_92d1e1eb} & 17.31 & 3.67 & 2917 & - \\ 



\midrule

\multirow{1}{*}{RTX 2060} &Animal-Spot~\cite{Bergler2022} & \textbf{280.73} & 21.23 & 4995 &   3491  \\  

& Whisper-tiny~\cite{whisper} & 273.57 & \textbf{106.81} & 4003 & \textbf{154} \\ 



& Whisper-base~\cite{whisper} & 137.83 & 54.66 & \textbf{3888} & 717 \\ 



& Whisper-large-v3~\cite{whisper} & 23.56 & 9.42 & 4228 & 4077 \\ 



& Wav2Vec 2~\cite{NEURIPS2020_92d1e1eb} & 78.73 & 32.04 & 3896 & 1077 \\ 



\bottomrule
\end{tabular}
\end{minipage}
\end{center}
\end{table*}

\subsection{Citizen Science Labelling Evaluation}

We compare the performance of non-expert labellers with that of an expert bioacoustician on the most likely audio files across 365 days at one site, and 31 days at another (396 samples total). The concordance between amateur and the expert labeller, as measured by Cohen's kappa, was 0.714, 0.580, and 0.229. The new instances of marine mammal data discovered in the paper have been labelled by the amateur labeller with a Cohen's Kappa of 0.714.

\subsection{Species and Ecotype Classification}

Species labels are available for 55,872 of 65,580 files. 392 are provided by ONC, 31,118 are provided by SanctSound~\cite{Hatch2024}, 22,567 are labelled by a human annotator in this project, and 1926 are pseudo-labelled by propagating labels to the files occurring immediately before or after a labelled event. These labels cover humpbacks (1501 h), orcas (2484 h), pacific white sided dolphins (246 h), sea lions (88 h), minke whales (0.3 h), fin whales (166 h), sperm whales (10 h), gray whales (0.2 h), or multiple species (20 h). These labels were also acquired using a combination of active learning and prioritising labelling likely orca files. During training, we omit the SanctSound labels, as they feature imprecise start and end times. A simple multiclass classifier is trained atop of the whisper-tiny embeddings. We evaluate the model on the DCLDE-2026 dataset (excluding ONC and OrcaSound contributions), which include humpback whales, orcas, and other dolphins. We map the pacific white-sided dolphin class to the other dolphin class in DCLDE-2026. The species-classifier obtains a top-1 classification accuracy of 53.2\% (11 train classes, 4 test classes).

Our dataset contains ecotype annotations for 2484 hours of orca data. This spans SRKWs (919 hours), Bigg's transients (230 hours), NRKW (3 hours), offshore (15 hours). SRKWs were labelled by pulsed calls. If the calls featured an S01, S03, S13, S16, S18, S19, S33, S36, S42, or S44, then the file was determined to be an SRKW~\cite{Ford_1984}. Bigg's were determined by T04 or T07 calls. NRKWs were more rare in the dataset, so there was no systematic labelling technique. Offshores were primarily determined by ruling out other ecotypes at deep water nodes. Pseudo-labels were generated for files neighbouring ecotype identifications, even if the pseudo-labelled file only contained echolocation or whistles.  Using these 4 classes, the ecotype-classifier obtains a top-1 accuracy of 33.6\% on the DCLDE-2026 dataset (4 train classes, 5 test classes). Our dataset does not include Southern Alaskan resident orcas.

\subsection{SRKW Encounters and Habitat Usage}
We compare the quantity of SRKW encounters with other sources of reporting defined by days with SRKW sightings(Table~\ref{tab:habitat}). While sightings have produced more events than this dataset, expanding the openly available hydrophone network to regions like Swiftsure Bank may enable this scalable approach to match those of sightings networks~\cite{Riera2019}. 

When comparing our observations to those in prior literature, we observe 170 additional whale-days~\cite{shields20232018}. This means that at least 20.8\% of SRKW whale-days were missed during the study's time period within the study's coordinates~\cite{shields20232018}. 68.9\% of these missed events occurred during daylight (solar altitude>0$^{\circ}$ ) 23.3\% occurred at night (solar altitude $\leq$ -12$^{\circ}$) with the remainder occurring at twilight. Over 60\% of the missed detections happened between July-September, with the missed events being evenly distributed across the years. Across our entire dataset, 47.2\% of vocalizations occurred during daylight, 38.8\% at night, and the remainder during twilight.

We also compare our observations to another prior work~\cite{https://doi.org/10.1111/mms.13012}. This work used 6 data sources, which involved contributed reports from hydrophones and sightings networks. We find roughly a tenth of the number of occurrences in the study region, likely due to a lack of hydrophone coverage. Despite this, the temporal trend of annual SRKW whale-days has a Pearson correlation of 69.1\%. Interestingly, the Bigg's population had a correlation coefficient of -25.6\%. One explanation could be that SRKWs are relatively noisy, so they are likely to be detected on the hydrophone each time they pass~\cite{doi:10.1098/rsos.231368, BARRETTLENNARD1996553}. Bigg's on the other hand, are silent while hunting~\cite{doi:10.1098/rsos.231368, BARRETTLENNARD1996553}.

\begin{table}[]
    \caption{Recorded southern resident killer whale events across a variety of sources through 2022. An "event" is defined as a day in which SRKWs are spotted. * 203.6 hours of confirmed events with the remainder being interpolated from the median event duration.}
    \label{tab:habitat}
    \centering
    \begin{tabular}{lrr}
        \toprule
        Source & Events & Audio (hrs)\\
        \midrule
         \citeauthor{shields20232018} (2018-2022)~\cite{shields20232018} & 647 & - \\
         PWWA (2018-2022) &  317 & - \\
         Orca Network & 1995 & - \\
         iNaturalist~\cite{inaturalist} & 14 & - \\
         OrcaSound & 122 & 214.6* \\ 
         ONC annotations & 10 & 3.1 \\
         DCLDE~\cite{Palmer2025} & 87  & 4.1 \\
         \midrule
         DORI (ours) & 560 & 919 \\
         \bottomrule
    \end{tabular}

\end{table}

\section{Discussion}

There are some easy tangible steps to produce better classifiers. In all of our experiments, we only use the whisper-tiny model with frozen weights during fine-tuning. This allows us to pre-compute and store all of the embeddings. Using a large whisper architecture or fine-tuning the entire model would likely improve performance. This model is designed with end-uses in mind. Bioacousticians may need to run these algorithms them aboard ships during field work, or the embeddings may be computed locally on the node to reduce the quantity of data that must be streamed. 

The algorithms we develop are specifically designed to detect pulsed calls and whistles within orca communication frequencies. By having a sampling rate of only 32 kHz, we exclude some dolphin species, while having a high-pass filter at 1 kHz we lose sensitivity for blue, sei, fin, humpback, gray, and north pacific right whales. Researchers interested in detecting these species should modify these parameters accordingly. 

Human labour availability is an important aspect of this work. Appropriate prior distributions are required to effectively conduct positive-unlabelled learning. Using traditional marine biologist surveys may reduce labelling burden when expanding to new species or new locations. Overall, the quantity of findings in these large archives using machine learning leaves ~\textit{more} work for marine biologists to do, rather than less. In addition, we observe that three of the amateur labellers had positive-valued Cohen's Kappa scores. We cite theoretical literature, and demonstrate empirically that the penalties to noisily labelled data are inconsequential. Datasets can feasibly be labelled by non-experts and still contribute meaningfully to science. Promising projects such as bioacoustic or satellite surveys for marine mammals~\cite{10.1371/journal.pone.0212532} may benefit from citizen science labelling initiatives (for example, ~\url{https://www.wwf.org.uk/learn/walrus-from-space}).

Unsupervised machine translation~\cite{lample2018word} has reignited interest in translating non-human animals. We note that this dataset is approximately the same size as the English librespeech corpus~\cite{7178964}. Unlike audiobooks, our dataset contains varying SNRs caused by ship noise, diarization challenges~\cite{mahon2025robust}, SRKW dialects~\cite{Ford_1984}, and contributions from nearly 100 individual orcas throughout the years. Nonetheless, we suspect that recent works investigating whale linguistics using this dataset may find similar properties with SRKWs. For example, humpback vocalizations follow the same statistical structure as human languages~\cite{doi:10.1126/science.adq7055}, sperm whales embed context in their vocalizations~\cite{Sharma2024}, and bottlenose dolphins will respond to whistles that individually identify them~\cite{dolphinnames}. The openly licensed dataset we curated should serve as a useful tool for investigating animal linguistics.

We welcome contributions from the community to add species, ecotype, pod, and individual identifications for this dataset through versioned merge requests.

\section{Conclusion}
We present the dataset for orca resident interpretation (DORI). This dataset is a curation of over 30 years of data resulting in 5298 hours of marine mammals and 919 hours of SRKW audio. We propose a novel active learning strategy that outperforms both entropy-based labelling and loss estimate labelling strategies. The newly-discovered quantity of data enables us to fine tune state of the art classifiers that outperform existing tools in terms of accuracy, speed, and energy efficiency. This work serves as a template for citizen science on vast uncurated datasets. These data are available for unbiased observations of SRKW behaviour, policy and planning activities for conservation, and to translate the SRKW dialect to human languages.

\section{Ethics, Fairness, and Limitations}
This paper focuses only on publicly accessible data through Ocean Networks Canada (CC-BY 4.0), OrcaSound (CC-BY-NC-SA 4.0), Ocean Observatories Initiative (publicly accessible under custom data usage agreement), and SanctSound~\cite{Hatch2024}. Our resulting curated dataset is publicly available, which reduces barriers to entry for marine biologists. Many of the hydrophones which data are collected from in Canada are located on Coast Salish waters off the coast of British Columbia. The coastal territories that coincide with the area of this study are mostly unceded and ancestral land of the Coast Salish. Hydrophone network operators, such as ONC use CARE principles for indigenous governance (https://www.gida-global.org/whoweare) in order to align with the UN's Declaration of Rights for Indigenous Peoples. While the source data is managed by partners, we respect indigenous rights to ownership, control, access, and possession, especially in unceded territorial waters. In some nations, whales are regarded as kin, and should be considered to have personhood and the afforded rights.

We intentionally omit data acquired from live-capture events and aquariums.

The Whisper models are distributed under the Apache-2.0 License. Our fine-tuned model weights and code are distributed under a BigScience Open RAIL-M License which restricts unethical use cases. These restrictions prohibit use for harming others, discriminating against others, impersonating others, or to be used in law enforcement circumstances, amongst others. 

Data collected in this study contains strong sampling bias, as hydrophones are placed in regions which are determined to be optimal and feasible. This may present as an over-abundance of marine mammal data compared to other locations within SRKW habitat. Passive acoustic monitoring overcomes some sampling bias from sightings and sea-based surveys by monitoring at night and during inclement weather.

\section{GenAI Disclosure}
Generative AI was used in the preparation of some code. It was also used to broaden the literature search beyond what was found by traditional indexed database searches. Generative AI was used in neither the writing of the paper, nor in the creation of citations, which were downloaded as \texttt{bibtex} or \texttt{RIS} format from source.

\begin{acks}
This project was funded by the UW Allen School
Postdoc Research Award.

Advanced computing resources are provided by the Digital Research Alliance of Canada (Alliance), the organization responsible for digital research infrastructure in Canada, and ACENET, the regional partner in Atlantic Canada. ACENET is funded by the Alliance, the provinces of New Brunswick, Newfoundland \& Labrador, Nova Scotia and Prince Edward Island, as well as the Atlantic Canada Opportunities Agency.

This material is based upon work supported by the National Science Foundation under Cooperative Agreement No. 1743430 (which supports the OOI) or other relevant NSF award number.

We would like to thank Mark Hamilton for providing the advice to use layer normalisation for the embeddings.
\end{acks}
\newpage

\bibliographystyle{ACM-Reference-Format}
\bibliography{sample-base}

\newpage
\appendix
\section{Appendix}

\centering
\begin{table*}
    \centering
    \begin{tabular}{lrrrrrr}
         \toprule
  Split & Lat. & Lon. & Date & Train & Valid. & Test \\
  \midrule
ONC-BACAX & 48.32 & -126.05 & 2009-2011 & \checkmark & & \\
ONC-BIIP & 49.30 & -123.11 & 2021 & \checkmark & & \\
ONC-CBCH.H1 & 47.76 & -127.73 & 2020-2021 & \checkmark & & \\
ONC-CBYIP & 69.11 & -105.06 & 2014-2016 & \checkmark & & \\
ONC-CQS64.H1 & 48.70 & -126.87 & 2017-2020 & \checkmark & & \\
ONC-CRIP & 50.02 & -125.24 & 2016-2021 & \checkmark & & \\
ONC-ECHO1.H1 & 49.04 & -123.32 & 2016 & \checkmark & & \\
ONC-ECHO1.H2 & 49.04 & -123.32 & 2016 & \checkmark & & \\
ONC-ECHO1.H3 & 49.04 & -123.32 & 2016 & \checkmark & & \\
ONC-ECHO2.H1 & 49.04 & -123.32 & 2014-2016 & \checkmark & & \\
ONC-ECHO2.H2 & 49.04 & -123.32 & 2018 & \checkmark & & \\
ONC-ECHO2.H3 & 49.04 & -123.32 & 2014-2015 & \checkmark & & \\
ONC-ECHO2.H4 & 49.04 & -123.32 & 2014-2016 & \checkmark & & \\
ONC-ECHO3.H1 & 49.04 & -123.32 & 2020-2021 & \checkmark & & \\
ONC-ECHO3.H2 & 49.04 & -123.32 & 2020-2021 & \checkmark & & \\
ONC-ECHO3.H3 & 49.04 & -123.32 & 2020-2021 & \checkmark & & \\
ONC-ECHO3.H4 & 49.04 & -123.32 & 2020-2021 & \checkmark & & \\
ONC-FAE & 49.08 & -123.34 & 2018 & \checkmark & & \\
ONC-FGPD & 48.81 & -125.28 & 2009-2021 & \checkmark & & \\
ONC-KEMO & 47.92 & -129.11 & 2016-2018 & \checkmark & & \\
ONC-LSBBL & 49.08 & -123.34 & 2016-2017 & \checkmark & & \\
ONC-LSHA.H1 & 49.08 & -123.34 & 2014 & \checkmark & & \\
ONC-LSHA.H2 & 49.08 & -123.34 & 2014-2015 & \checkmark & & \\
ONC-LSHA.H3 & 49.08 & -123.34 & 2014-2015 & \checkmark & & \\
ONC-LSHA.H4 & 49.08 & -123.34 & 2014 & \checkmark & & \\
ONC-LSHA.H5 & 49.08 & -123.34 & 2014 & \checkmark & & \\
ONC-NC27.H1 & 47.76 & -127.76 & 2014-2015 & \checkmark & & \\
ONC-NCBC & 48.43 & -126.18 & 2010-2011 & \checkmark & & \\
ONC-PVIP & 48.66 & -123.49 & 2019-2021 & \checkmark & & \\
ONC-SCVIP & 49.04 & -123.43 & 2015-2021 & \checkmark & & \\
ONC-SEHA.H5 & 49.04 & -123.32 & 2014-2015 & \checkmark & & \\
ONC-SEVIP & 49.04 & -123.32 & 2019-2020 & \checkmark & & \\
ONC-SGC.H2 & 49.04 & -123.426 & 2017-2018 & \checkmark & & \\
ONC-SGE.H1 & 49.04 & -123.32 & 2008-2015 & \checkmark & & \\
ONC-SGE.H2 & 49.04 & -123.32 & 2011-2012 & \checkmark & & \\
ONC-USDDL & 49.08 & -123.33 & 2021 & \checkmark & & \\
ONC-USSLP & 49.085 & -123.330 & 2018-2019 & 
\checkmark & & \\

OOI-CE04OSBP & 44.37 & -125.00 & 2018-2021  & \checkmark & & \\
OOI-CE02SHBP & 44.64 & -124.31 & 2015-2016 & \checkmark & & \\
OOI-RS01SLBS & 44.52 & -125.39 & 2016-2020  & \checkmark & & \\
OOI-RS01SBPS & 44.53 & -125.39 & 2016-2021 & \checkmark & & \\

SanctSound-OC01 & 48.39 & -124.65 & 2019-2020 & \checkmark & & \\
SanctSound-OC02 & 48.49 & -125.00 & 2019-2021 & \checkmark & & \\
SanctSound-OC03 & 47.32 & -124.41 & 2020-2021 & \checkmark & & \\
SanctSound-OC04 & 48.04 & -124.73 & 2019-2020 & \checkmark & & \\
SanctSound-NRS & 47.76& -125.52 & 2014-2020 & \checkmark & & \\

\bottomrule
    \end{tabular}
    \caption{Training splits by location and time}
    \label{tab:training}
\end{table*}

\newpage

\centering
\begin{table*}
    \centering
    \begin{tabular}{lrrrrrr}
         \toprule
  Split & Lat. & Lon. & Date & Train & Valid. & Test \\

\midrule
ONC-CBCH.H1 & 47.76 & -127.73 & 2022 & & \checkmark & \\
ONC-CRIP & 50.02 & -125.24 & 2022 & & \checkmark & \\
ONC-ECHO3.H1 & 49.04 & -123.32 & 2022 & & \checkmark & \\
ONC-ECHO3.H2 & 49.04 & -123.32 & 2022 & & \checkmark & \\
ONC-ECHO3.H3 & 49.04 & -123.32 & 2022 & & \checkmark & \\
ONC-ECHO3.H4 & 49.04 & -123.32 & 2022 & & \checkmark & \\
ONC-FGPD & 48.81 & -125.28 & 2022 & & \checkmark & \\
ONC-PVIP & 48.66 & -123.49 & 2022 & & \checkmark & \\
ONC-SCVIP & 49.04 & -123.43 & 2022 & & \checkmark & \\
ONC-USDDL & 49.08 & -123.33 & 2022 & & \checkmark & \\
OOI-CE04OSBP & 44.37 & -125.00 & 2022 & & \checkmark & \\
SanctSound-OC02 & 48.49 & -125.00 & 2022 & & \checkmark & \\
\midrule
ONC-FGPD  & & & 2023 & & & \checkmark \\
ONC-PSGCH  & & & 2023/08 & & & \checkmark \\
DCLDE2026-DFO-CRP & & &  & & & \checkmark \\
DCLDE2026-DFO-WDLP & 48.72 & -125.1 & 2022 & & & \checkmark \\
DCLDE2026-DFO-WDLP & 49.2 & -123.3 & 2021 & & & \checkmark \\
DCLDE2026-DFO-WDLP & 48.94 & -123.4 & 2021 & & & \checkmark \\
DCLDE2026-DFO-WDLP & 48.8 & -123.4 &  2021 & & & \checkmark \\

DCLDE2026-Scripps & 47.4 & -124.7 & 2008-2012 & & & \checkmark \\
DCLDE2026-Scripps & 47.5 & -125.4 & 2011-2013 & & & \checkmark \\
DCLDE2026-SIMRES & & &  & & & \checkmark \\
DCLDE2026-SMRU & & &  & & & \checkmark \\
DCLDE2026-UAF & 60.3 & -147.0 &  & & & \checkmark \\
DCLDE2026-UAF & 59.9 & -151.8 &  & & & \checkmark \\
DCLDE2026-UAF & 60.2 & -147.8 &  & & & \checkmark \\
DCLDE2026-UAF & 59.7 & -149.5 &  & & & \checkmark \\
DCLDE2026-VFPA & - & - &  & & & \checkmark \\
DeepAl & & & 2017-2018 & & & \checkmark \\
\bottomrule
    \end{tabular}
    \caption{Validation and test splits by location and time}
    \label{tab:valtest}
\end{table*}

\begin{table*}[h]
\begin{center}
\begin{minipage}{\textwidth}
\caption{Cetacean presence/absence classification specificity at 95\% sensitivity on several held out test sets. Expressed as a percentage.}\label{tab:spec}%

\centering
\begin{tabular}{lrrrr}
\toprule
  Model & Deepal & DCLDE-2026~\cite{Palmer2025} & FGPD & PSGCH.H3 \\
   & ComParE~\cite{orcaspot}  & excl. ONC \& OS & 2023 & 2023/08 \\
\midrule

LDA\footnote{Algorithm implemented in python to operate across platforms. Hyper-parameter tuning was conducted to outperform the PAMGuard defaults.} & 5.26 & 3.52 & 2.44 & 9.52 \\
ROCCA\footnote{Algorithm implemented in python to operate across platforms. Hyper-parameter tuning was conducted to outperform the PAMGuard defaults.} ~\cite{rocca} & 5.39 & 3.42\footnote{Generated contours were randomly subsampled so that only 1000 contours were selected}  & 7.32 & 19.05 \\
ANIMAL-SPOT~\cite{Bergler2022} & \textbf{19.48} & 3.77 & 2.44 & 0 \\
Whisper-tiny~\cite{whisper} + LR & 4.99 & 9.07 & 10.24 & 14.29 \\
Whisper-tiny~\cite{whisper} + SGD & 10.55 & 4.16 & 12.20 & 38.10 \\
Whisper-tiny~\cite{whisper} + FT & 7.28 & \textbf{18.6} & 9.76 & 76.19 \\
Whisper-base~\cite{whisper} + FT & 18.47 & 4.16 & 6.83 & 28.57 \\
Whisper-large-v3~\cite{whisper} + FT & 15.55 & 8.86 & \textbf{12.68} & \textbf{85.71} \\
Wav2Vec 2~\cite{NEURIPS2020_92d1e1eb} + FT & 8.88 & 4.41 & 7.80 & 19.05 \\
\bottomrule
\end{tabular}
\end{minipage}
\end{center}
\end{table*}

\begin{table*}[h]
\begin{center}
\begin{minipage}{\textwidth}
\caption{Cetacean presence/absence classification AUPRC several held out test sets.}\label{tab:AUPRC}%

\centering
\begin{tabular}{lrrrr}
\toprule
  Model & Deepal & DCLDE-2026~\cite{Palmer2025} & FGPD & PSGCH.H3 \\
   & ComParE~\cite{orcaspot}  & excl. ONC \& OS & 2023 & 2023/08 \\
\midrule
LDA\footnote{Algorithm implemented in python to operate across platforms. Hyper-parameter tuning was conducted to outperform the PAMGuard defaults.}  & 0.205 & 0.791 & 0.437 & 0.353 \\
ROCCA\footnote{Algorithm implemented in python to operate across platforms. Hyper-parameter tuning was conducted to outperform the PAMGuard defaults.} ~\cite{rocca} & 0.186 & 0.777\footnote{Generated contours were randomly subsampled so that only 1000 contours were selected}  & 0.472 & 0.375 \\
ANIMAL-SPOT~\cite{Bergler2022} & 0.379 & 0.765 & 0.450 & 0.282 \\
Whisper-tiny~\cite{whisper} + LR & 0.210 & 0.846 & 0.432 & 0.495 \\
Whisper-tiny~\cite{whisper} + SGD & 0.277 & 0.850 & 0.586 & 0.761 \\
Whisper-tiny~\cite{whisper} + FT & 0.232 & \textbf{0.923} & \textbf{0.666} &\textbf{ 0.919} \\
Whisper-base~\cite{whisper} + FT & \textbf{0.421} & 0.913 & 0.589 & 0.659 \\
Whisper-large-v3~\cite{whisper} + FT & 0.367 & 0.922 & 0.626 & 0.807 \\
Wav2Vec 2~\cite{NEURIPS2020_92d1e1eb} + FT& 0.309 & 0.871 & 0.517 & 0.566 \\
\bottomrule
\end{tabular}
\end{minipage}
\end{center}
\end{table*}

\begin{figure*}[h]
  \centering
  \includegraphics[width=\linewidth]{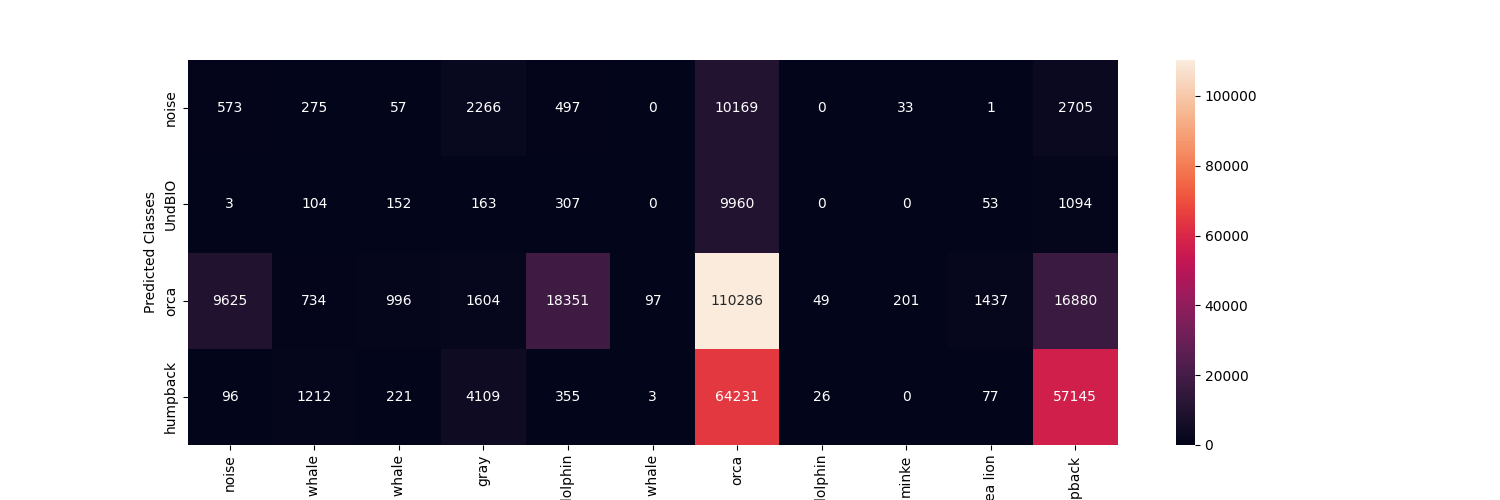}
  \caption{Confusion matrix for the actual classes from DCLDE 2026 vs. the whisper-tiny species classifications. }
  \label{fig:species_heatmap}
  \Description{A heatmap.}
\end{figure*}

\begin{figure*}[h]
  \centering
  \includegraphics[width=\linewidth]{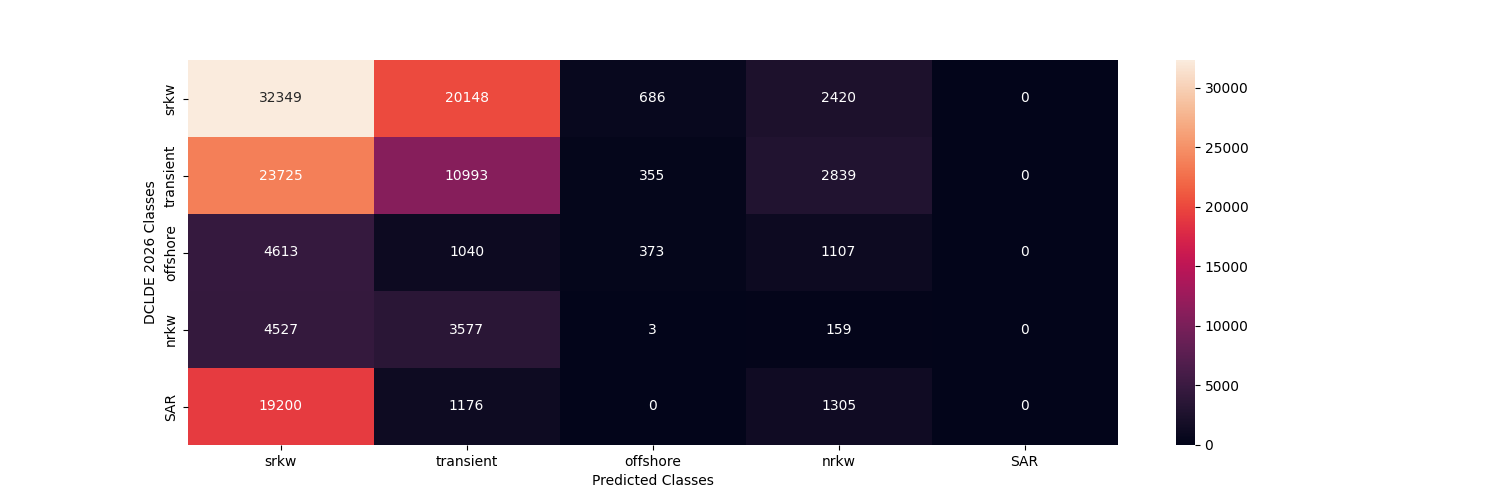}
  \caption{Confusion matrix for the actual ecotype classes from DCLDE 2026 vs. the whisper-tiny ecotype classifications. }
  \label{fig:ecotype_heatmap}
  \Description{A heatmap.}
\end{figure*}

\begin{figure*}[h]
    \centering
    
      \includegraphics[width=1\linewidth]{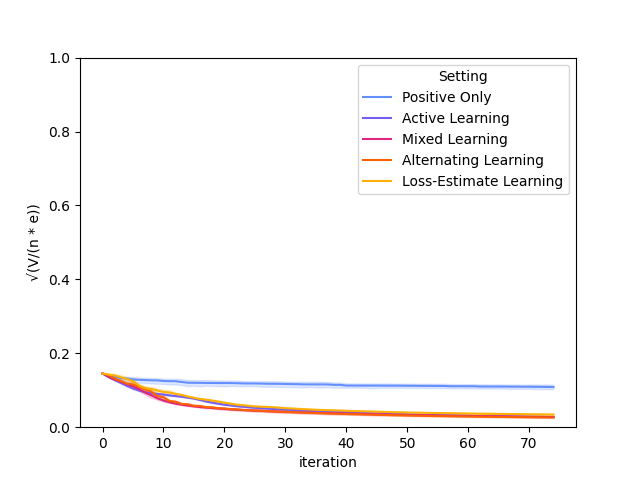}
      \caption{The criterion to determine which convergence regime is applicable.}
      \label{fig:converge} 
    
    \Description{The criterion to determine which convergence regime is applicable.}

\end{figure*}

\begin{figure*}[ht]
    \centering
    
    \begin{subfigure}[b]{0.7\linewidth}
      \includegraphics[width=0.98\linewidth]{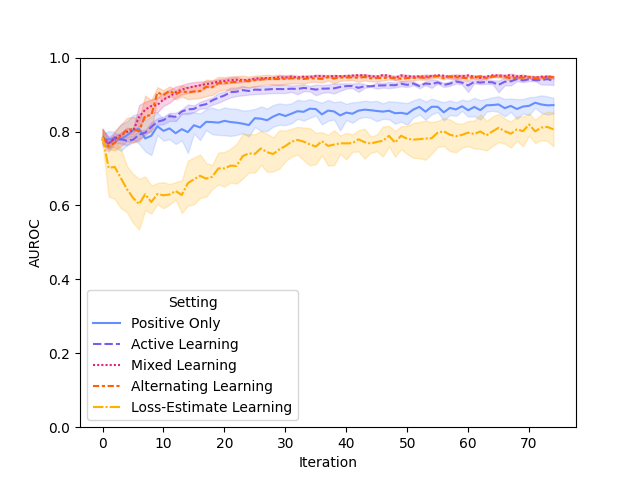}
      \caption{AUROC vs active learning iterations with no label noise}
      \label{fig:auroca} 
    \end{subfigure}
    
    \medskip 
    \begin{subfigure}[b]{0.7\linewidth}
      \includegraphics[width=0.98\linewidth]{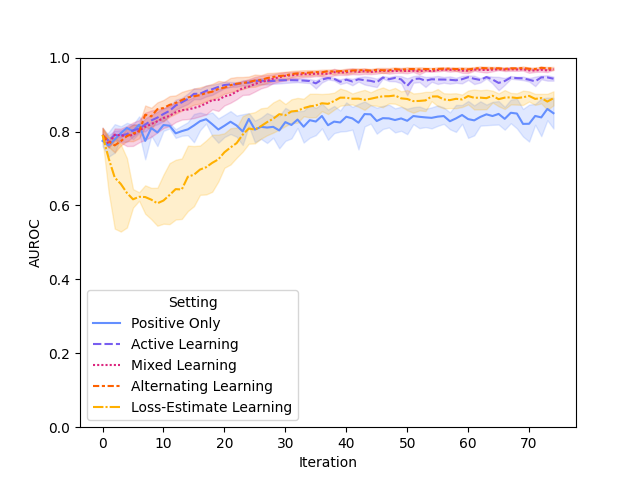}
      \caption{AUROC vs. active learning iteration with label noise.}
      \label{fig:aurocb}
    \end{subfigure}
    
    \caption[AUROCNoise]{%
    AUROC across active learning iterations for a variety of settings}
    \Description{Active learning experiments under label noise}

\end{figure*}

\begin{figure*}[ht]
    \centering
    
    \begin{subfigure}[b]{0.7\linewidth}
      \includegraphics[width=0.98\linewidth]{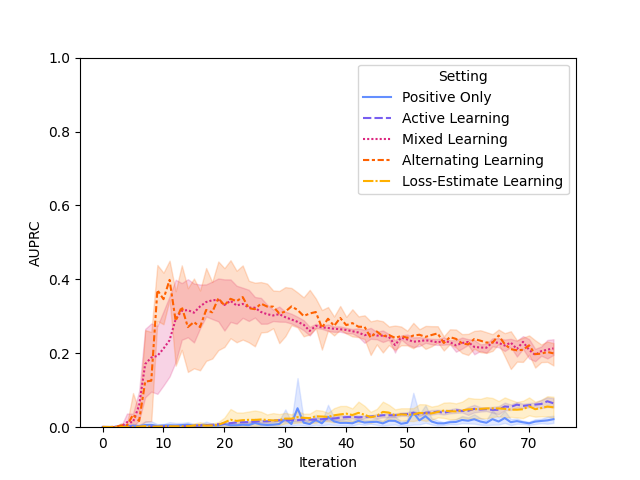}
      \caption{AUPRC vs active learning iterations with no label noise}
      \label{fig:auprca} 
    \end{subfigure}
    
    \medskip 
    \begin{subfigure}[b]{0.7\linewidth}
      \includegraphics[width=0.98\linewidth]{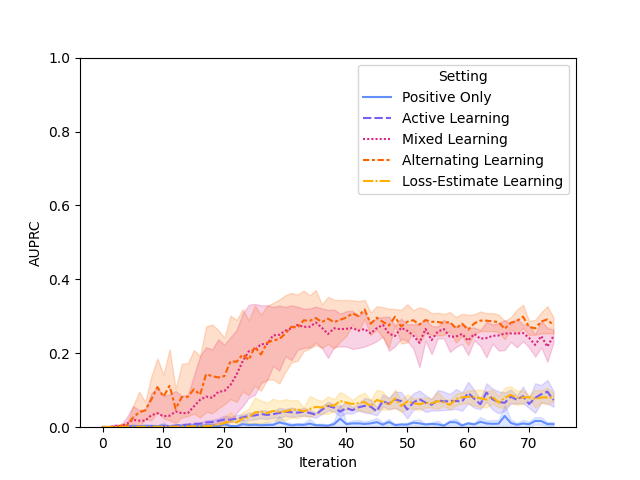}
      \caption{AUPRC vs. active learning iteration with label noise.}
      \label{fig:auprcb}
    \end{subfigure}
    
    \caption[AUPRCNoise]{%
    AUPRC across active learning iterations for a variety of settings}
    \Description{Active learning experiments under label noise}

\end{figure*}

\end{document}